# Risk-aware Vehicle Motion Planning Using Bayesian LSTM-Based Model Predictive Control

Yufei Huang, and Mohsen Jafari

***Abstract*—Understanding the probabilistic traffic environment is a vital challenge for the motion planning of autonomous vehicles. To make feasible control decisions, forecasting future trajectories of adjacent cars is essential for intelligent vehicles to assess potential conflicts and react to reduce the risk. This paper first introduces a Bayesian Long Short-term Memory (BLSTM) model to learn human drivers' behaviors and habits from their historical trajectory data. The model predicts the probability distribution of surrounding vehicles' positions, which are used to estimate dynamic conflict risks. Next, a hybrid automaton is built to model the basic motions of a car, and the conflict risks are assessed for real-time state-space transitions based on environmental information. Finally, a BLSTM-based Model Predictive Control (MPC) is built to navigate vehicles through safe paths with the least predicted conflict risk. By merging BLSTM with MPC, the designed neural-based MPC overcomes the defect that traditional MPC is hard to model uncertain conflict risks. The simulation results show that our proposed BLSTM-based MPC performs better than human drivers because it can foresee potential conflicts and take action to avoid them.**

## I. Introduction

Precise and timely vehicle maneuver and position recognition are vital for Advanced Driver Assistance Systems (ADAS) and Automated Driving Systems (ADS). The timely understanding of the behaviors of other road users helps these systems assess collision risks and proactively avoid traffic conflicts. However, partial observability of the road environment and unexpected driver behaviors would significantly reduce the accuracy of trajectory prediction [1]. In recent years, Vehicle-to-infrastructure (V2I) communications have been studied to capture real-time data and provide road users with advisories to enhance driving safety [2]. Roadside LiDARs and cameras surveil road segments, and traffic information can be extracted from video images or cloud points [3]. Traffic data are then processed and analyzed on cloud servers, and warning messages are sent to drivers' smart devices to avoid potential near misses [4]. Taking advantage of the V2I framework, decision-making models can be built to plan a safe trajectory for intelligent vehicles. Vehicle trajectories can be represented as a state sequence that combines paths and maneuvers [5]. Vehicle motion planning helps vehicles perform maneuvers, such as slowing down or turning left, at specific locations to avoid obstacles on the road. For a nonlinear and high-dimensional urban road environment, safe motion planning is a complex problem that requires vehicles to react to real-time environmental changes.

This paper proposes a probabilistic vehicle trajectory prediction model that considers the uncertain behaviors of human drivers. A merged Bayesian Neural Network (BNN) and a Long Short-Term Memory (LSTM) [6] model is proposed to predict stochastic short-term vehicle positions. The probabilistic distribution of forecasted vehicle locations is obtained using Monte Carlo (MC) Dropout technique [7], and is used to compute traffic conflict risks. Stochastic conflict risk for each vehicle is generated using the predicted position distribution, which helps autonomous vehicles and human drivers evaluate the road environment and make control decisions. Also, the paper develops a control model that combines physics-based dynamics and stochastic conflict risk predictions. The Bayesian LSTM (BLSTM) is embedded into a Model Predictive Control (MPC) for vehicle motion planning and control. A hybrid automaton describes basic vehicle motions. The proposed neural network-based MPC can reduce the predicted nonlinear conflict risk and guide the vehicle through a safe corridor. This work is experimentally supported by real-world traffic data collected from roadside cameras in a downtown area in New Brunswick, New Jersey, USA. The data is processed to obtain vehicle information such as GPS coordinates, speed, and lane geometry. The paper's contributions are three folds: (i) A BLSTM model that learns from historical vehicle trajectory data and estimates the uncertainty of vehicle positions in the future. (ii) A dynamic and probabilistic conflict risk distribution that can be utilized for vehicle motion planning and safety warning. (iii) A hybrid control model that navigates a vehicle through a predicted safe corridor. The approach combines BLSTM with traditional MPC to take advantage of model-based control and model-free learning.

This paper is organized as follows. Section II presents an overview of probability vehicle trajectory prediction methods. Path planning and motion control techniques are also briefly reviewed in this section. Section III introduces the Bayesian LSTM deep learning structure for predicting stochastic vehicle locations and conflict risks. Section IV illustrates our Bayesian LSTM-based Model Predictive Control model to guide the vehicle following a real-time computed path that has minimized collision risk with surrounding vehicles. Section V shows the model training process and the validation results of

Y. Huang, and M. Jafari are with the Department of Industrial and Systems Engineering, Rutgers University – New Brunswick, Piscataway, NJ 08854, USA (e-mail: yh639@scarletmail.rutgers.edu; jafari@soe.rutgers.edu).

our proposed method. Section VI is the summary and provides some future thoughts.

## II. Related Works

Trajectory prediction of an adjacent vehicle is a significant subject in situational awareness and evaluating potentially hazardous conditions in advance [8]. Much research attention has been paid to this area, including deterministic trajectory prediction algorithms [9] and uncertainty estimation of vehicle paths [10]. Probabilistic vehicle trajectory forecasting models are commonly established with Bayesian Networks (BN). To predict a stochastic vehicle path, Bayesian deep learning models are trained with prior knowledge, such as road geometry and different driving maneuvers [11]. Besides BN, other methods like Hidden Markov Model (HMM) [12] and fuzzy logic [13] have also been used for probabilistic prediction. The advantage of BN is that it can combine different sources of knowledge and explicit treatment of uncertainty and support for decision analysis [14]. [15] proposed an architecture that uses Bayesian networks to detect qualitative maneuvers, and [16] used Dynamic Bayesian Network for interactive vehicle maneuver prediction and classification of pre-defined actions. Researchers have also explored using Recurrent Neural Networks (RNNs) to solve sequential trajectory prediction problems [17]. [18] combines a Bayesian Neural Network (BNN) with RNN for probabilistic vehicle trajectory prediction. The BNN outputs trajectory distributions based on the environment, and the RNN ensures the feasibility of the distributions according to physical models. Real-time traffic conflict detection is vital for early-stage collision warnings. [15] predicts vehicle path and estimates conflict risks using the Time-to-Collision (TTC) metric. Some other conflict indices include the deceleration rate to avoid collision (DRAC) [19], hourly conflict risk index (HCRI) [20], and severity index (SI) [21]. These safety indices focus on vehicle information, historical crash record, and collision severity to measure the conflict risk. Most of these methods use a set threshold to distinguish unsafe conditions. There isn't a continuous risk score to quantify the risk level.

Controlling an intelligent vehicle is sometimes solved by accomplishing two sub-tasks: path planning and tracking. One well-known path-planning approach is the Artificial Potential Field (APF), which originated from obstacle avoidance in robotics [22]. Considering conflict risks as a potential field, a safe corridor can be generated for an intelligent vehicle. [23] brought up a novel idea that combines APF with Model Predictive Control (MPC) for controlling automated vehicles (AVs). However, MPC is a model-based controller that highly depends on the model accuracy, and it's hard to address stochastic risk [24]. The behavior decomposition method solves motion planning by breaking it into several independent behavior units, such as collision avoidance and tracking. [25] divided the navigation behaviors into goal-reaching and collision avoidance. The complexity of the learning process of the navigation task is reduced to separate sub-tasks. Other well-known motion planning methods include but are not limited to the dynamic window approach (DWA) [26]. DWA samples the environment at the current time step to obtain the robot's action state at the next moment. [27] combined DWA with the A* algorithm [28] and achieved both local obstacle avoidance and global path planning. Compared to the model-based motion planning methods mentioned above, model-free controllers, such as type-2 fuzzy logic control [29] or Artificial Neural Network (ANN) [30], works better if the mathematical model of the system is hard to build [31]. [32] proposed a Q-learning-based [33] path-planning method that utilizes cased-based reasoning [34] to learn past experiences. Their method increases the convergence efficiency of the original Q-learning algorithms in solving motion planning problems. However, these model-free control models rely on massive data and are expensive to train [35]. Neural network (NN)-based MPC has been proposed to overcome the limitations of a model-based controller by using a NN to describe the system, which was previously hard to interpret [36]. These hybrid controllers can exploit the advantages of both model-free and model-based control.

There are significant gaps between theoretical advances and real-world practices in vehicle motion planning and control: (i) Many existing vehicle motion planning algorithms lack the ability to consider the probabilistic behaviors of other road users. Their estimated conflict zones sometimes have a static size and don't change with aggressive or defensive driving behaviors; (ii) The uncertainty and nonlinearity of vehicle trajectories make it hard to implement traditional model-based control methods and physics-based vehicle controllers [37]. The model-free vehicle control models are black boxes and are not interpretable. They lack knowledge about the road environment and are highly dependent on the training data [38].

## III. Probabilistic Vehicle Trajectory Prediction

In this section, a probabilistic vehicle trajectory prediction architecture is introduced. The structure combines BNN with LSTM, enabling time-series forecasting and uncertainty estimation. The Bayesian LSTM learns prior knowledge based on road geometry and road traffic information. It predicts the possible location for each vehicle within a confidence level. Conflict risk is then defined based on the predicted position distribution. The next sections provide more details about our technical approach. In the sequel, Boldface is used to indicate vectors.

### *A. BNN with MC Dropout*

Given a set of $M$ inputs $\boldsymbol{X} = \{\boldsymbol{i_1}, \boldsymbol{i_2}, \dots, \boldsymbol{i_m}\}$ and a set of $M$ outputs $\boldsymbol{y} = \{\boldsymbol{o_1}, \boldsymbol{o_2}, \dots, \boldsymbol{o_m}\}$, a feed-forward neural network $\boldsymbol{y} = f^{\boldsymbol{w}}(\boldsymbol{X})$ conveys information through each input $\boldsymbol{X}$ to corresponding output $\boldsymbol{y}$ via the connections of one or more hidden layers. Note that $i_j$ and $o_j$ , $j \in [1 \quad 2 \quad \dots \quad m]$, don't have to be the same dimension. These connections are associated with weights $\boldsymbol{w}$ and activation functions. The training process is to estimate parameters $\widehat{\boldsymbol{w}}$ that makes predicted $\widehat{\boldsymbol{y}}$ close to the actual value $\boldsymbol{y}$. Different from a standard NN, which learns the parameter values $\widehat{\boldsymbol{w}}$, a Bayesian

Neural Network (BNN) seeks the conditional distribution of the weights given to the training data. As shown in Figure 1, a traditional NN has deterministic weights after training, while a BNN's weights follow a normal distribution. $h^{(1)}$ in the figure refers to the hidden layer. The objective of a Bayesian neural network is to increase the accuracy of the inferencing. The BNN models the uncertainty by quantifying the predicted output $\hat{\boldsymbol{y}}$ with standard error $\eta$ and $\alpha$ level confidence interval, defined by $[\hat{\boldsymbol{y}} - Z_{\alpha/2}\boldsymbol{\eta} \quad \hat{\boldsymbol{y}} + Z_{\alpha/2}\boldsymbol{\eta}]$, where $Z_{\alpha/2}$ represents the $\alpha/2$ quantile of a standard Normal distribution.

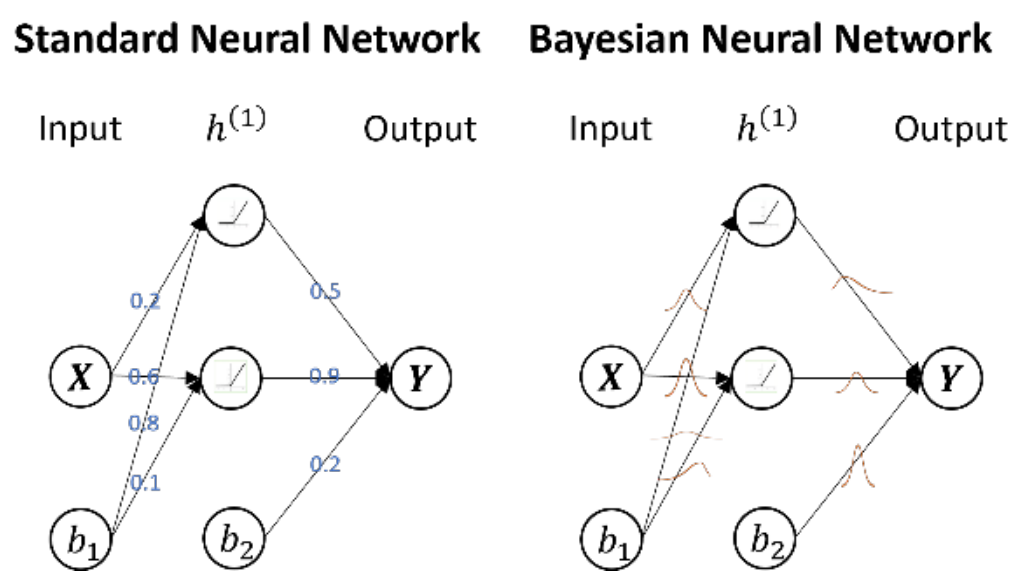

**Fig. 1.** Comparison between a standard NN and BNN.

Here, $p(\hat{\boldsymbol{w}}|\boldsymbol{X},\boldsymbol{y})$ is the posterior knowledge or Bayesian inference. Monte Carlo dropout (MC dropout) [7] is commonly applied to calculate $p(\hat{\boldsymbol{w}}|\boldsymbol{X},\boldsymbol{y})$ The MC dropout works as follows: given the inputs $\boldsymbol{X}$ and the outputs $\boldsymbol{y}$, we randomly dropout a neuron in each layer with probability $p$, where $p$ follows Bernoulli distribution $p\sim\text{Bernoulli}(p_d)$. In this work, $p_d = 0.5$ so that each neuron has the same probability of being removed or remaining in the neural network during each training step, as is shown in Figure 2. After the BNN finishes training, a new input $\boldsymbol{x}^*$ is fed into the model repeatedly $k$ times and the neurons keep the same dropout rate $p_d$ to performe the predictions. We will have $k$ outputs $\hat{\boldsymbol{y}}^* = \{\hat{\boldsymbol{y}}_1^*, \hat{\boldsymbol{y}}_2^*, \dots, \hat{\boldsymbol{y}}_k^*$ with mean $\mu = \frac{1}{k}\sum_{m=1}^{k}\hat{\boldsymbol{y}}_m^*$ and variance $\sigma^2 = \frac{1}{k}\sum_{m=1}^{k}(\hat{\boldsymbol{y}}_m^* - \mu)^2$. In this work, $\alpha = 6$, $i.e.$, $[\mu - 3\sigma \quad \mu + 3\sigma]$, which gives us a 95% confidence interval for $\hat{y}^*$:

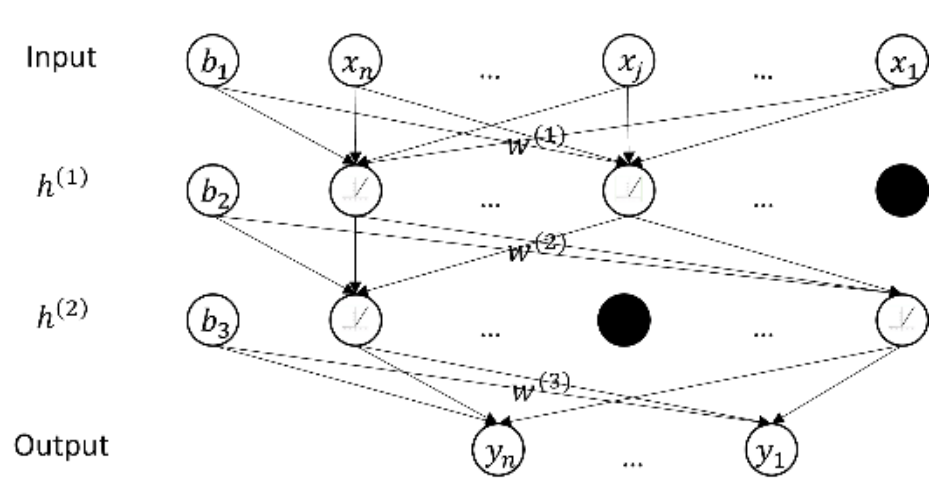

**Fig. 2.** BNN with MC dropout.

*B. Bayesian LSTM for Vehicle Trajectory Prediction*

Long Short-Term Memory (LSTM) is a Recurrent Neural Network (RNN) widely used for sequence prediction problems because it can learn long-term dependencies. To generate uncertainty of time-series forecasting, Bayesian LSTM is applied together with a BNN. The structure of Bayesian LSTM is shown in Figure 3. The sequential data first passes through the standard LSTM cell. Then the output of LSTM is fed into a BNN, which gives us the final prediction result.

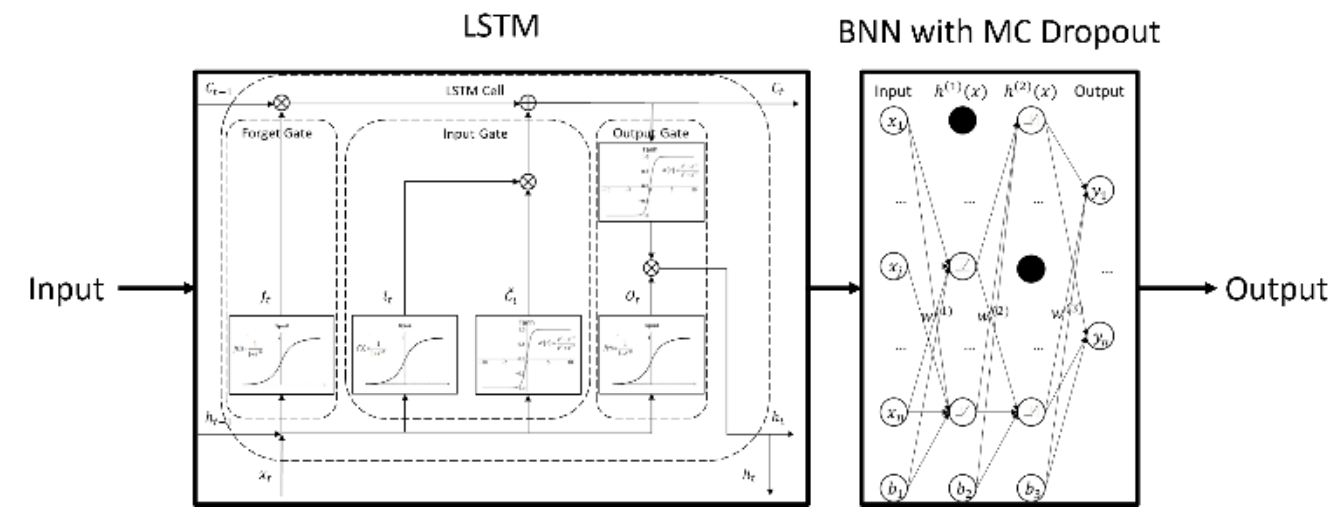

**Fig. 3.** Bayesian LSTM structure.

To use the proposed BLSTM architecture for probabilistic vehicle trajectory prediction, real-time vehicle driving behaviors are obtained from roadside cameras using a longitudinal scanline-based vehicle trace reconstruction method [39]. The technique is a computer vision algorithm that extracts vehicle profiles from high-angle traffic video for all vehicles on a road segment frame by frame. It combines scanline-based trajectory extraction and feature-matching coordinate transformation to detect vehicles and get their coordinates through time. The extracted vehicle profile for each vehicle in a frame contains 23 features listed in Appendix A. The outputs of the BLSTM model are the predicted coordinates for the vehicle in the next time step, denoted by $(x, y)$. By repeating the prediction $k$ times, the mean and standard deviation of the outputs can be calculated; these are denoted by $\mu_x$ and $\mu_y$ for means, and $\sigma_x$ and $\sigma_y$ for standard deviations.

Figure 4 shows the workflow of the BLSTM model. For each vehicle, its profile in the current time step is concatenated with its historical information in the previous two time steps and passed into the model. The BNN used in this paper contains one hidden layer and ten neurons. The model outputs are the vehicle's predicted longitude and latitude in the next step. The input or output dimension $[a \quad \times \quad b]$ in the plot means that it is a matrix that has $a$ rows and $b$ columns.

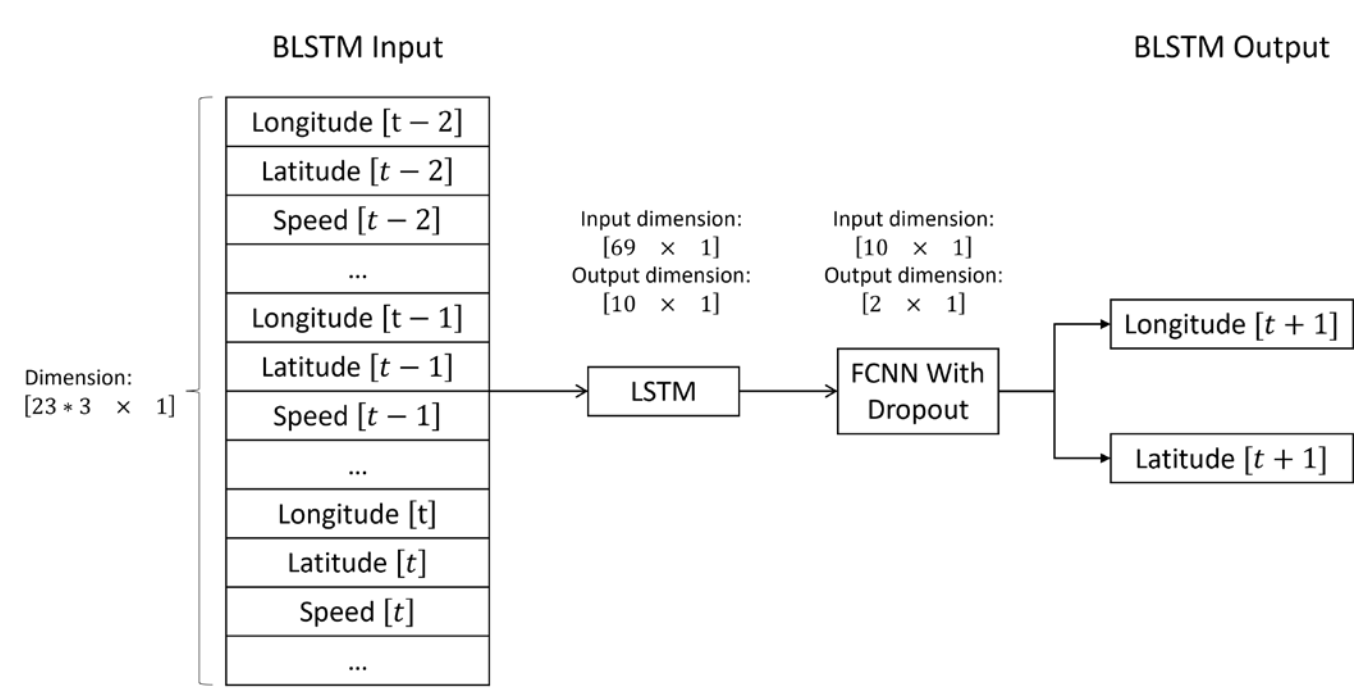

**Fig. 4.** BLSTM structure.

The pre-trained BLSTM is a general probabilistic trajectory prediction model. The parameters in the pre-trained model can be re-trained to better capture driving behaviors for new vehicles entering the road segment. The re-training procedure is illustrated in Figure 5. When the pre-trained model cannot perform well to fit vehicles on the road (e.g., the prediction accuracy decreases), the fine-tuned model can be built by re-training the model using recently collected vehicle trajectory data. The data can be split into a small re-training dataset and a large testing dataset. For example, the first 2 seconds of each new trajectory containing 40 frames are used to update the model parameters. Then, the fine-tuned model can be tested and used to predict the future trace of the vehicles.

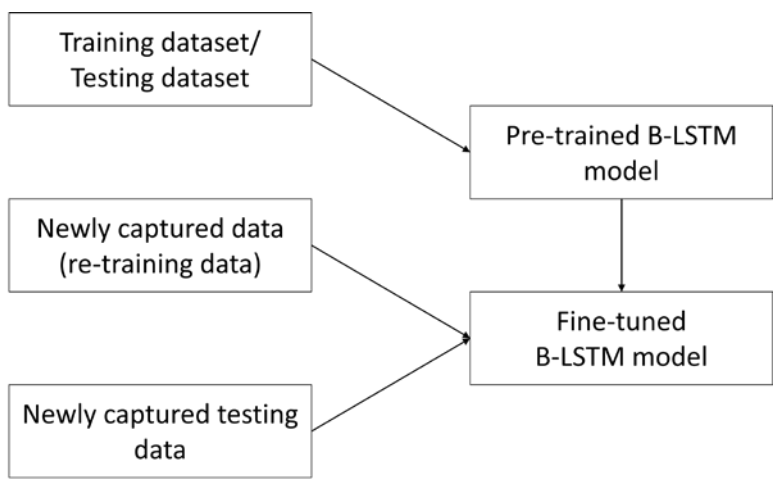

**Fig. 5.** Re-training process for new cars in the road segment.

### *C. Conflict Risk*

The BLSTM framework provides the predicted mean and standard deviation of each vehicle in the next time step. A Bivariate Normal Distribution with pdf

$$p(\mathrm{x},\mathrm{y}) = \frac{e^{\left(-\frac{1}{2}(\boldsymbol{x}-\boldsymbol{\mu})^T\Sigma^{-1}(\boldsymbol{x}-\boldsymbol{\mu})\right)}}{2\pi\sqrt{|\Sigma|}}$$

defines the potential location of individual vehicles, where mean is $\boldsymbol{\mu} = \begin{bmatrix}\mu_x \\ \mu_y\end{bmatrix}$, and the variance-covariance matrix is $\Sigma = \begin{bmatrix}\sigma_x^2 & 0 \\ 0 & \sigma_y^2\end{bmatrix}$. Given the PDF for each vehicle at time $t$, the conflict risk function created by each vehicle at a certain location $(x, y)$ is defined as:

$$f_r = \frac{p(x,y)}{p(\mu_x, \mu_y)}$$

$f_r$ is a normalized distribution on the predicted mean since the risk score is set to be one at the center of the PDF. These conflict risks create potential conflict zones for each vehicle, as shown in Figure 6 for a case study with three vehicles on the road. The road centerline and both road boulders are also assigned a risk score equal to one. In Figure 6, the risks assigned to the road centerline and road shoulder in the same direction are removed for a clear perspective.

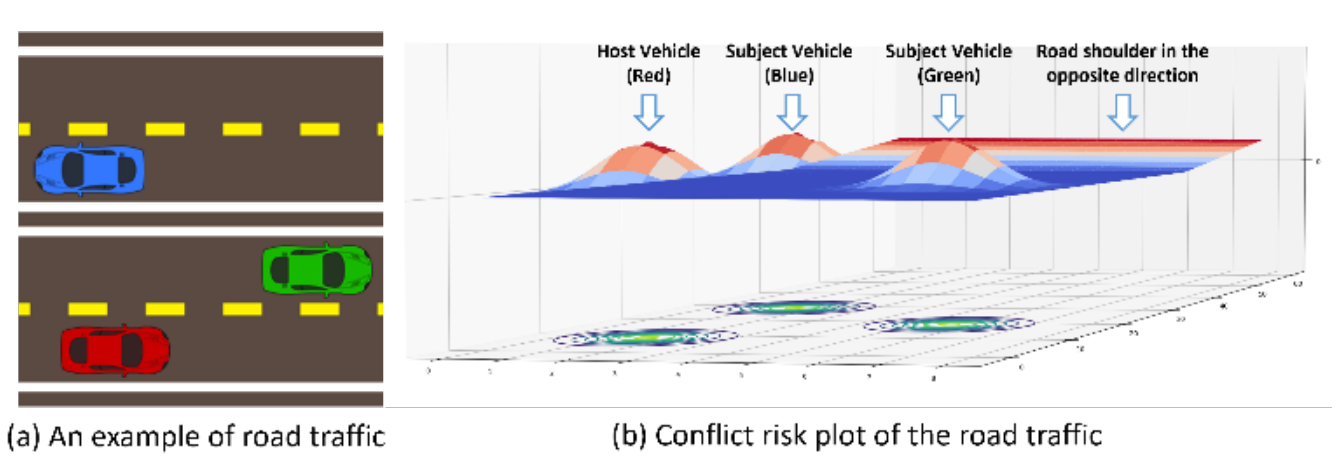

**Fig. 6.** Conflict zones created by BLSTM.

Figure 7 shows an example of a potential conflict between two vehicles, which are referred to as host $h$ and subject $s$. $r_t^h$ represents the conflict risk of the host vehicle at time $t$, due to the subject vehicle. It is defined by:

$$r_t^h = f_r^s(x_t^h, y_t^h)$$

Where $f_r^s$ is the conflict risk function created by the subject vehicle. $(x_t^h, y_t^h)$ is the position of the host vehicle at time $t$.

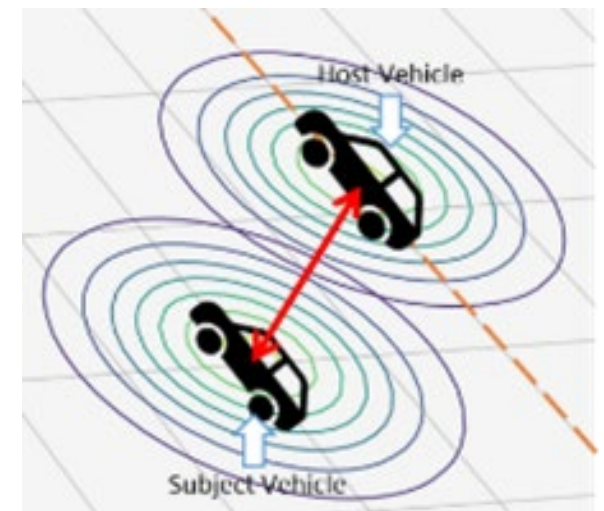

**Fig. 7.** An example of a potential conflict.

## IV. Model-Based Vehicle Motion Control

This section first illustrates a simplified hybrid automaton that models the basic motions of a vehicle. Then, the BLSTM-based Model Predictive Control framework is established to scheme a vehicle's movement and manipulate the car to reduce conflict risks. The probabilistic vehicle trajectory prediction described in Section III is used as a dynamic safety constraint for the neural MPC. The MPC continuously solves the optimal control problem by looking for a state-space transition over a pre-defined time horizon that minimizes future risk.

### *A. Hybrid Automaton for Vehicle Motion Planning*

In this study, a hybrid automaton model is built to model both continuous states of vehicles' motions and discrete maneuver transitions between the states. The automaton contains five states, which are: (i) cruise at current speed, (ii) change lane to the right, (iii) change lane to the left, (iv) accelerate, and (v) decelerate. All the states are fully connected and recurrent, as shown in Figure 8. The state-space transition decisions are made at each time step.

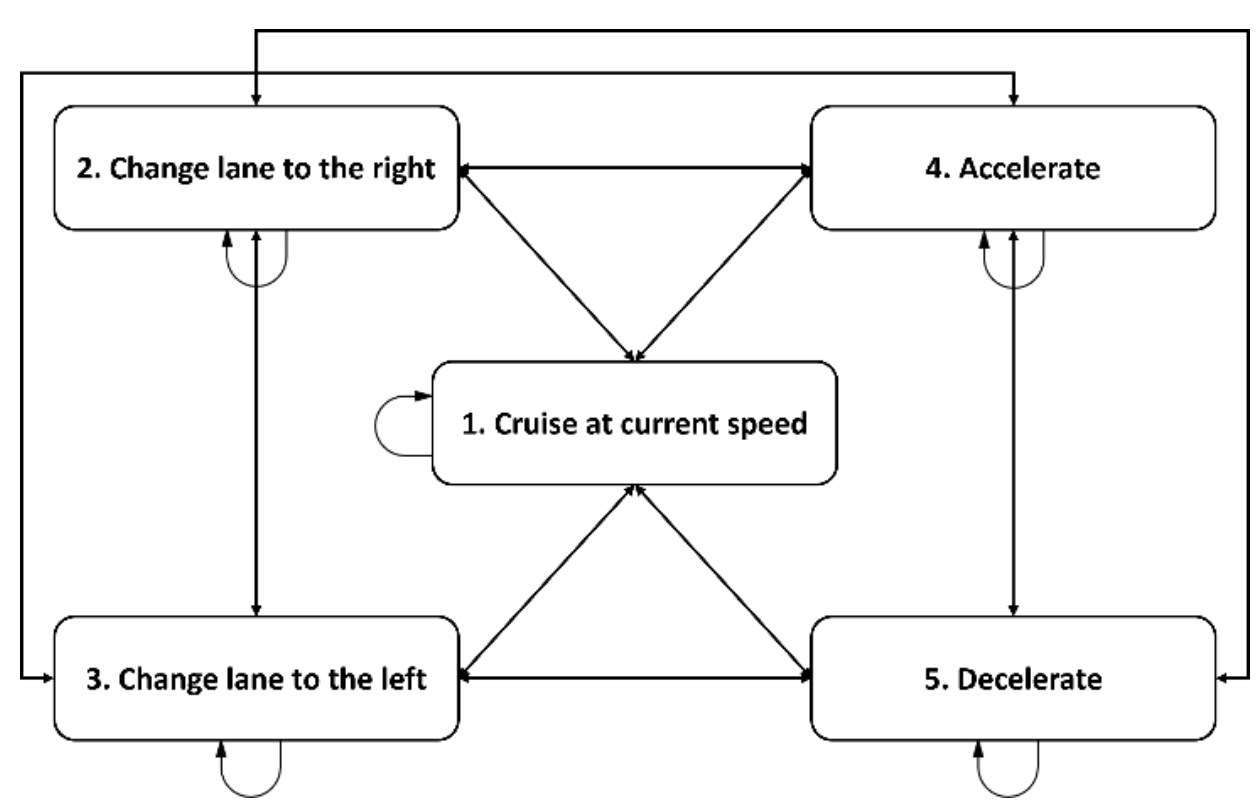

**Fig. 8.** A hybrid automaton model for vehicle control.

As illustrated in Figure 9, the position of a vehicle is defined by the middle point of the front bumper of a car. Let the vehicle's coordinates be $(x, y)$, measured in meters. Use $v$ to represent the vehicle's speed (in meters per second) and $a$ as the acceleration or deceleration rate for the car (in meters per squared second). The vehicle accelerates when $a$ is positive and decelerates when $a$ is negative. The yaw angle of the vehicle is denoted by $\delta$, which shows the degrees by which the vehicle turns to the left or right. Use $[\ldots \quad t-1 \quad t \quad t+1 \quad \ldots]$ to denote time steps, where a time step is one second and $\Delta t = t+1-t = 1$. The study done by [40] indicates that the average lane change angular speed for human drivers is 8 degrees per second. To simplify the problem, we assume a constant $\delta = 8$. Also, [41] suggests that the safe and comfortable acceleration rate is $2\,m/s^2$. The recommended deceleration rate is 3.4 $m/s^2$. Therefore, the acceleration rate and deceleration rate are set as fixed values to simplify the model.

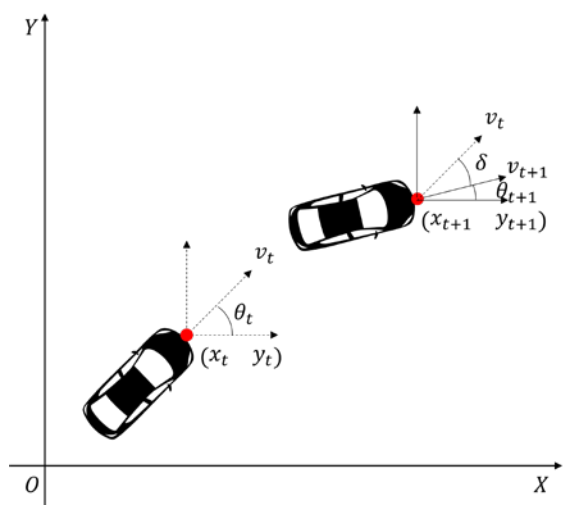

**Fig. 9.** A hybrid automaton model for vehicle control.

The dynamic equations for the vehicle in each state can be modeled as follows:

1) Cruise at the current speed

$$a_t = 0$$
$$v_{t+1} = v_t$$
$$\theta_{t+1} = \theta_t$$
$$x_{t+1} = x_t + v_t \cdot \cos\theta_t$$
$$y_{t+1} = y_t + v_t \cdot \sin\theta_t$$

2) Change lane to the right

$$a_t = 0$$
$$v_{t+1} = v_t$$
$$\theta_{t+1} = \theta_t + \delta$$
$$x_{t+1} = x_t + v_t \cdot \cos\theta_{t+1}$$
$$y_{t+1} = y_t + v_t \cdot \sin\theta_{t+1}$$

3) Change lane to the left

$$a_t = 0$$
$$v_{t+1} = v_t$$
$$\theta_{t+1} = \theta_t - \delta$$
$$x_{t+1} = x_t + v_t \cdot \cos\theta_{t+1}$$
$$y_{t+1} = y_t + v_t \cdot \sin\theta_{t+1}$$

4) Accelerating

$$a_t = 2$$
$$v_{t+1} = v_t + a_t \cdot \Delta t$$
$$\theta_{t+1} = \theta_t$$
$$x_{t+1} = x_t + v_t \cdot \cos\theta_t + \frac{1}{2} a_t \cdot \cos\theta_t$$
$$y_{t+1} = y_t + v_t \cdot \sin\theta_t + \frac{1}{2} a_t \cdot \sin\theta_t$$

5) Decelerating

$$a_t = -3.4$$
$$v_{t+1} = v_t + a_t \cdot \Delta t$$
$$\theta_{t+1} = \theta_t$$
$$x_{t+1} = x_t + v_t \cdot \cos\theta_t + \frac{1}{2} a_t \cdot \cos\theta_t$$
$$y_{t+1} = y_t + v_t \cdot \sin\theta_t + \frac{1}{2} a_t \cdot \sin\theta_t$$

At each time step, the vehicle needs to decide whether it will maintain its current state or make state transitions. A controller is required to determine the long-term conflict risk for each vehicle transition and choose the action that can lead the vehicle through the safest path.

### *B. BLSTM-based Model Predictive Control*

The basic idea of MPC is to use dynamic models to predict the future behaviors of the controlled system and keep optimizing the constrained problem in a receding horizon. In this work, the MPC is combined with the BLSTM model explained in section III to search for a safe path in the long run and apply the control maneuvers defined in Section IV-A to follow the path.

Denote the conflict risk for our host vehicle $h$ at time step $t$ as $R_t^h$. We want to minimize the summation of the $R_t^h$ over the future $N$ time steps. The objective function can be written as Equation (1).

$$\min \sum_{n=1}^{N} R_{t+n}^h \tag{1}$$

In this paper, we set $N = 5$. $R_t^h$ is determined by the predicted conflict risks created by all other surrounding vehicles at the host vehicle's position, which are written as $r_t^{h-}(x_t^h, y_t^h)$). The maximum value defines the risk of the host vehicle among all conflict risks. An assumption is made that the host vehicle chooses an action $u_t^h$ every time step $t$. $u_t^h$ is discrete and can take one of the five states defined in the hybrid automaton in Section IV-A. The constraints of the MPC can be defined as:

$$\boldsymbol{P}_t^h = \begin{bmatrix} \boldsymbol{l}_t^h \\ \boldsymbol{s}_t^h \\ \boldsymbol{d}_t^h \end{bmatrix} \tag{2}$$

$$\boldsymbol{P}_{t+1}^h = f_d\,(\boldsymbol{P}_t^h, u_t^h) \tag{3}$$

$$r_{t+1}^{h-} = f_r(\boldsymbol{P}_t^{h-}, \boldsymbol{P}_{t-1}^{h-}, \boldsymbol{P}_{t-2}^{h-}) \tag{4}$$

$$R_t^h = \max(r_t^{h-}(x_t^h, y_t^h)) \tag{5}$$

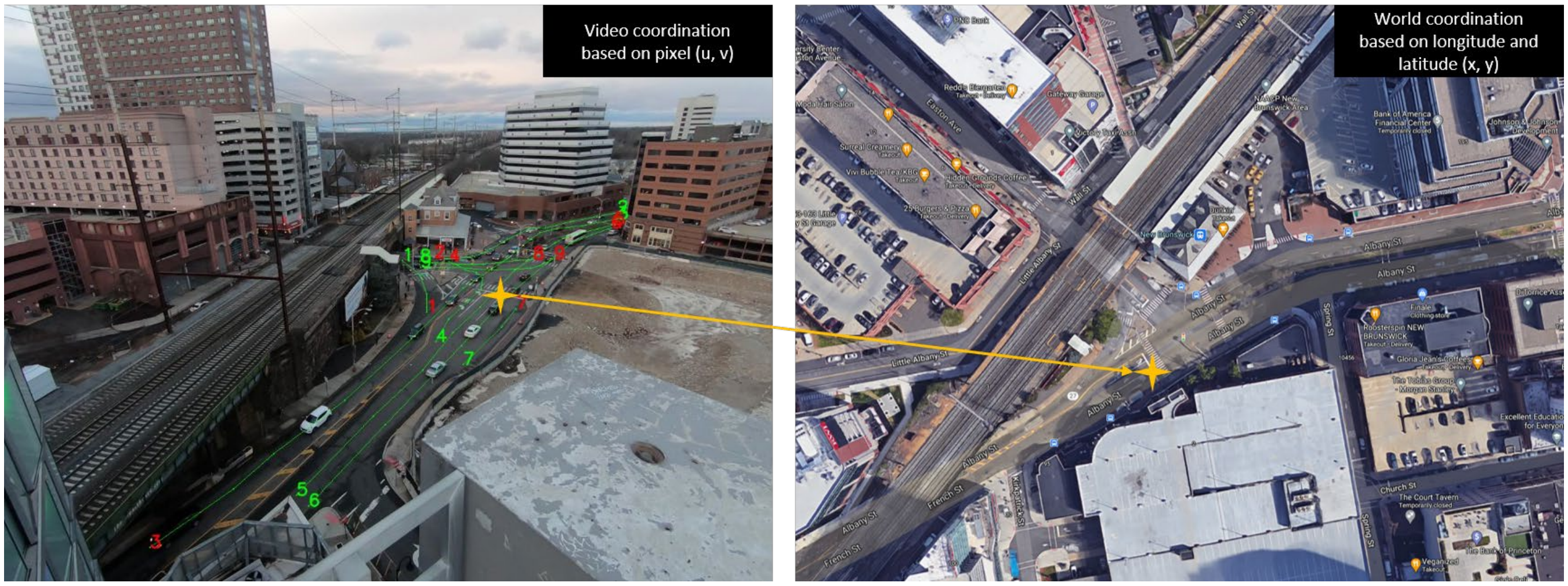

**Fig. 11.** Scope of the test bed for data collection.

At each time step, the vehicle profile $\boldsymbol{p}_t$, containing vehicle location information $\boldsymbol{l}_t^h$, speed information $\boldsymbol{s}_t^h$ and distance information $\boldsymbol{d}_t^h$, are obtained using roadside cameras as defined in Equation (2). $\boldsymbol{l}_t^h$ is a vector that contains the $x$ and $y$ coordinates of the vehicle. $\boldsymbol{s}_t^h$ consists of (i) speed, (ii) acceleration/ deceleration rate, (iii) speed difference to the front vehicle, (iv) acceleration difference to the front vehicle, (v) the speed of the front vehicle, and (vi) the acceleration/ deceleration of the front vehicle. $\boldsymbol{d}_t^h$ includes (i) headway distance, (ii) direction, (iii) whether the vehicle is in Queue, and (iv) geometry of the lane (straight, left turn, right turn, or merge). In Equation (3), $\boldsymbol{P}_t^{h-}$ indicates the vehicle profiles without the host vehicle at time $t$. In Equation (4), $f_d$ follows the dynamic equations of Section IV-A and calculates the state information of the host vehicle in the next time step given control action $u_t^h$. $f_r$ represents the BLSTM model described in Section III. Given the vehicle profile of one vehicle in the previous three time steps, $f_r$ predicts the conflict risk created by each vehicle in the next time step as defined in Section III-C. The conflict risk for the entire road segment $R_t^h$ is then defined by the union of the risk distributions of all other vehicles at the time step $t$ as is shown in Equation (5). The maximum conflict risk value created by surrounding vehicles is selected as the host vehicle's risk score.

The BLSTM-MPC can be solved by Depth-First Search (DFS), as is shown in Figure 10. Starting from the current state, the BLSTM-MPC calculates the risks for each possible action and keeps doing the calculation until $t = t + N$. If an action is infeasible (e.g., the action will run the vehicle off the roadway), the controller prunes the entire branch and searches for the other arms of the tree. The controller picks the branch with the least summation of risks among all feasible branches as the optimal solution. Then, the action at $t = t + 1$ from the selected branch is chosen to be the current control action, and the rest of the actions are dropped. The controller makes the decision repeatedly at each time step.

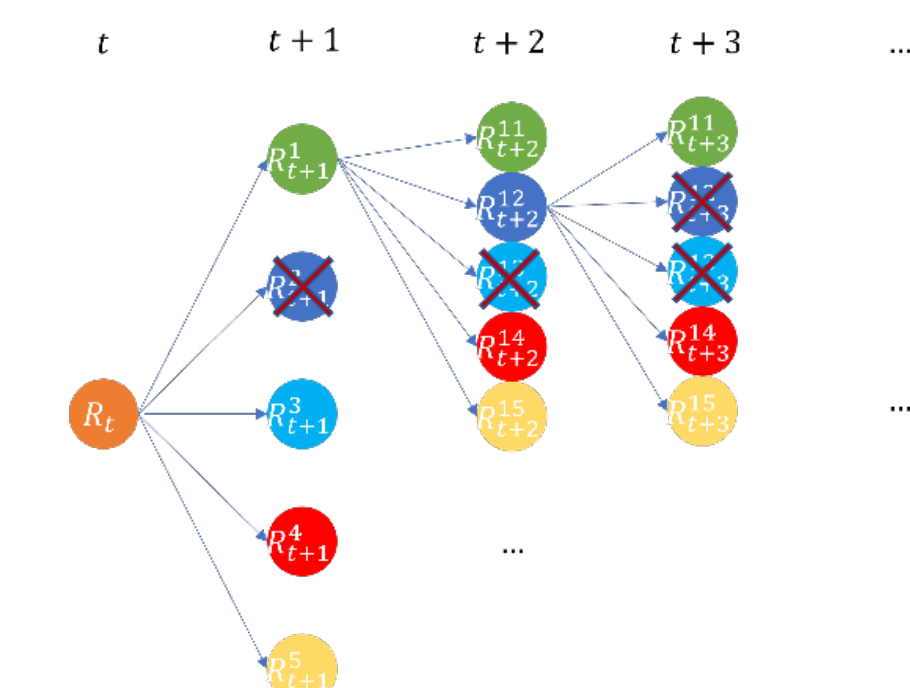

**Fig. 10.** A hybrid automaton model for vehicle control.

## V. Model Training and Validation

This section exhibits testing results for BLSTM based on actual traffic data. Also, the use cases for conflict risk prediction are illustrated, and the simulations for controlling an intelligent vehicle using BLSTM-MPC are presented

### *A. Data Collection*

Real-world traffic data are collected at the intersection of Easton Avenue and Albany Street in New Brunswick, New Jersey, USA. The roadside camera is placed on the roof of a public garage. It provides a bird's eye view of the intersection, as shown in Fig. 11. The position of each vehicle in a video frame is mapped to the GPS coordinates using the longitudinal scanline-based vehicle trace reconstruction method [39]. The data collection was performed at 3:35 PM on May 16th, 2021, and lasted 40 minutes. One thousand eight hundred vehicle trajectories were successfully detected and extracted for training and testing the BLSTM model. The dataset is publicly available at [42].

80% of the vehicles were used for training and testing the model. The remaining 20% of the vehicles were excluded from the training and testing process. They were used to test the model fine-tuning for new cars in the road segment. For the model fine-tuning data, the first 40 frames of each vehicle are used to fine-tune the pre-trained model, and the rest are used to test the model's accuracy.

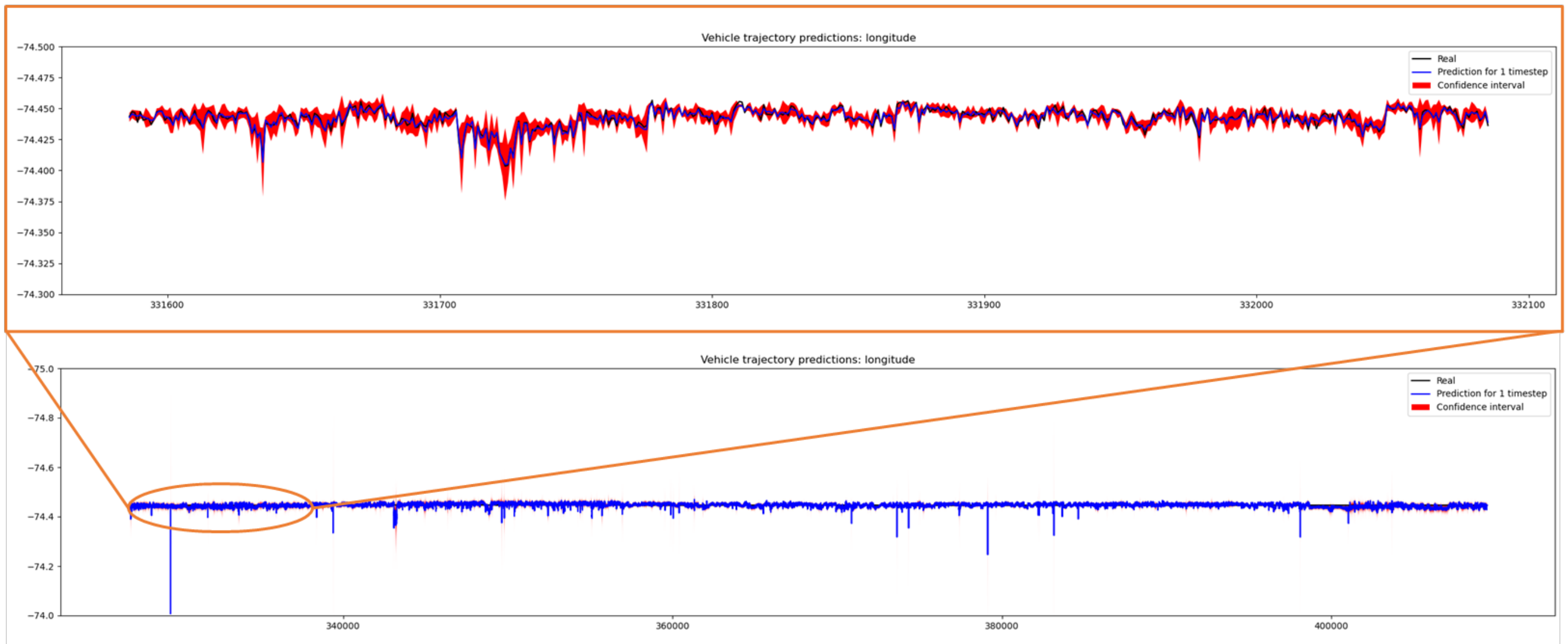

(a) BLSTM prediction results for longitude

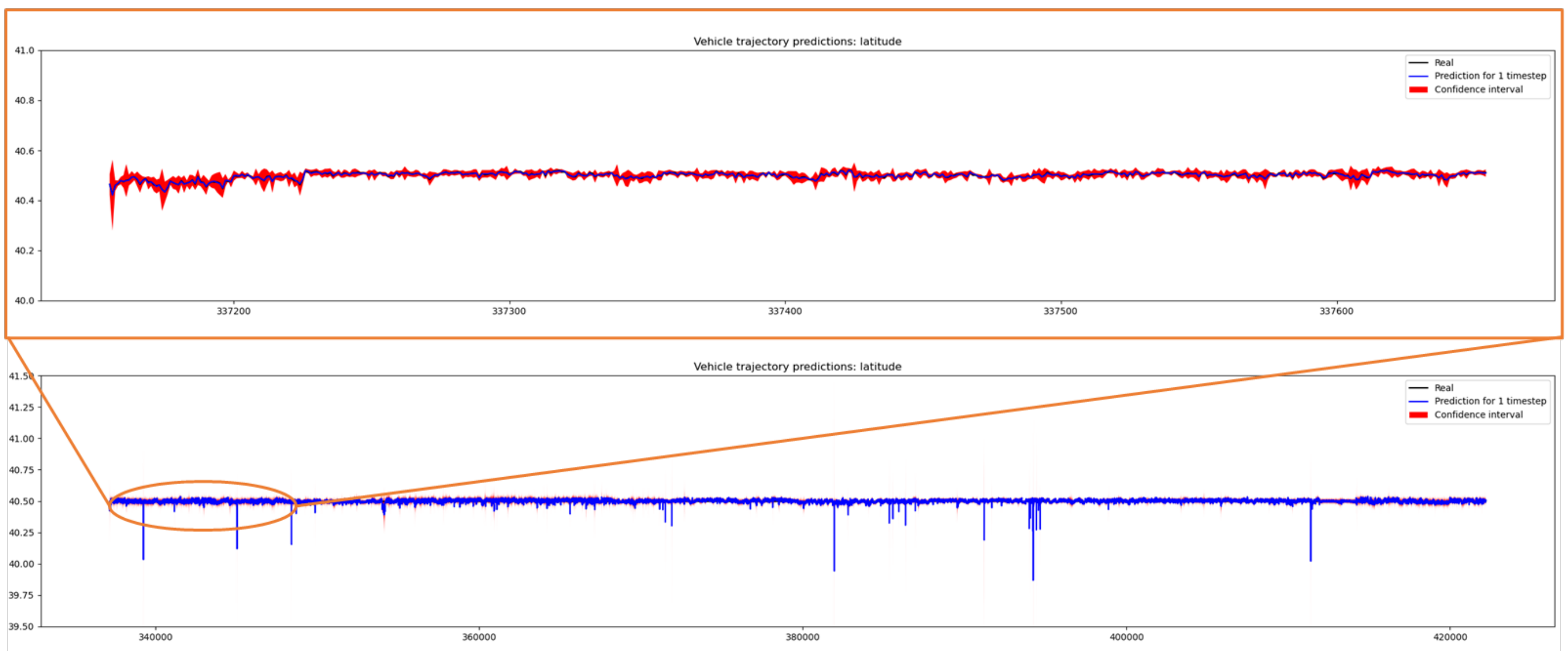

(b) BLSTM prediction results for latitude

**Fig. 12.** BLSTM testing results.

### *B. Validation Results for BLSTM*

Figure 12 shows us the testing results for the BLSTM model. The predicted trace of longitude and latitude values for each vehicle are concatenated one after another for illustration purposes. We use the vehicle information in the previous three frames to predict the possible range of longitude and latitude in the next frame for each vehicle. 98.08% of the predicted longitude values and 98.62% of the predicted latitude values are within the 95% confidence interval. Figure 13 shows the fine-tuning results, where 91.66% of the predicted longitude values and 89.45% of the predicted latitude values are within the 95% confidence interval.

An experiment is designed to test the impact on predictions of the length of time lags. The same BLSTM network structure is used to predict the longitude and latitude values with different window sizes. Figure 14 indicates that using the previous two time steps along with the current time step gives the best prediction results. The reason is that if the lookback period for trajectory prediction is too long, the network will be more complex, and we need to increase the number of neurons or layers in the BNN. In this case, the training time will be increased, and more effort must be put into tuning the network's hyper-parameters. Therefore, we set the window size to three, which has a good enough forecasting performance given the current setup.

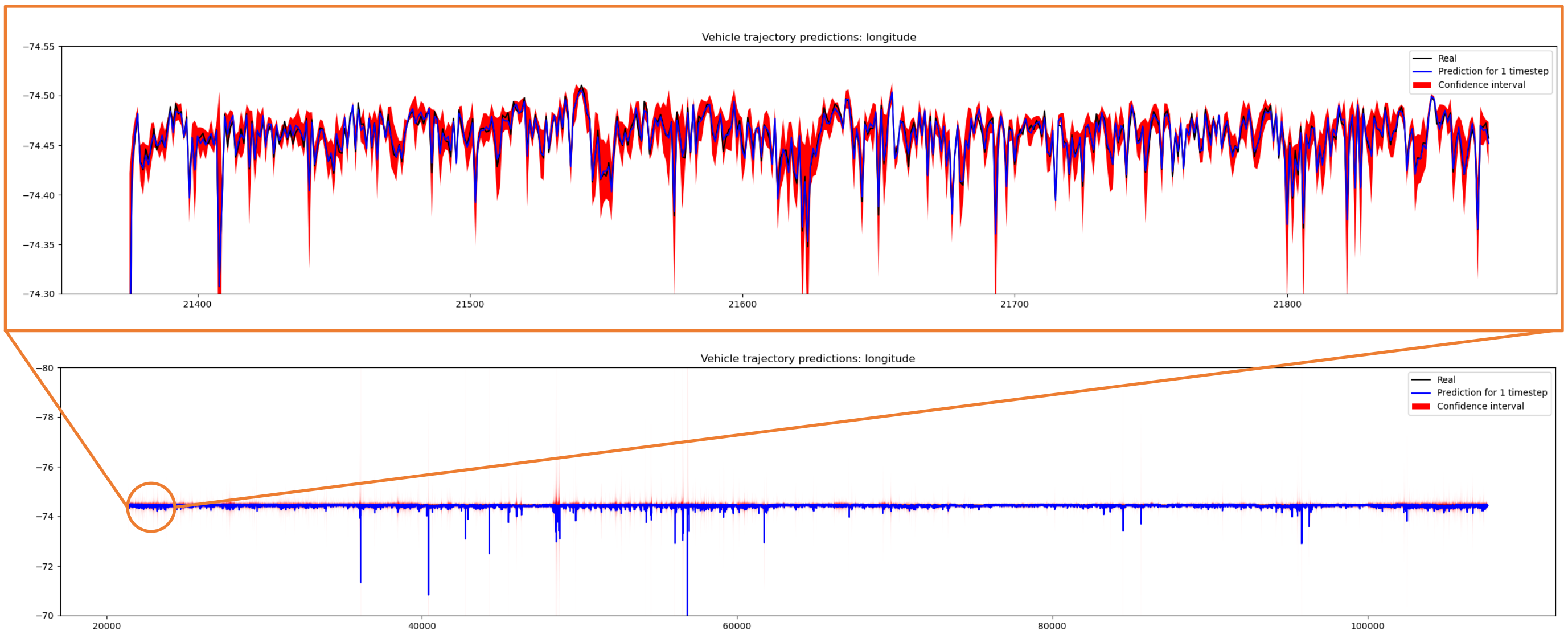

(a) Fine-tuned BLSTM model for longitude prediction

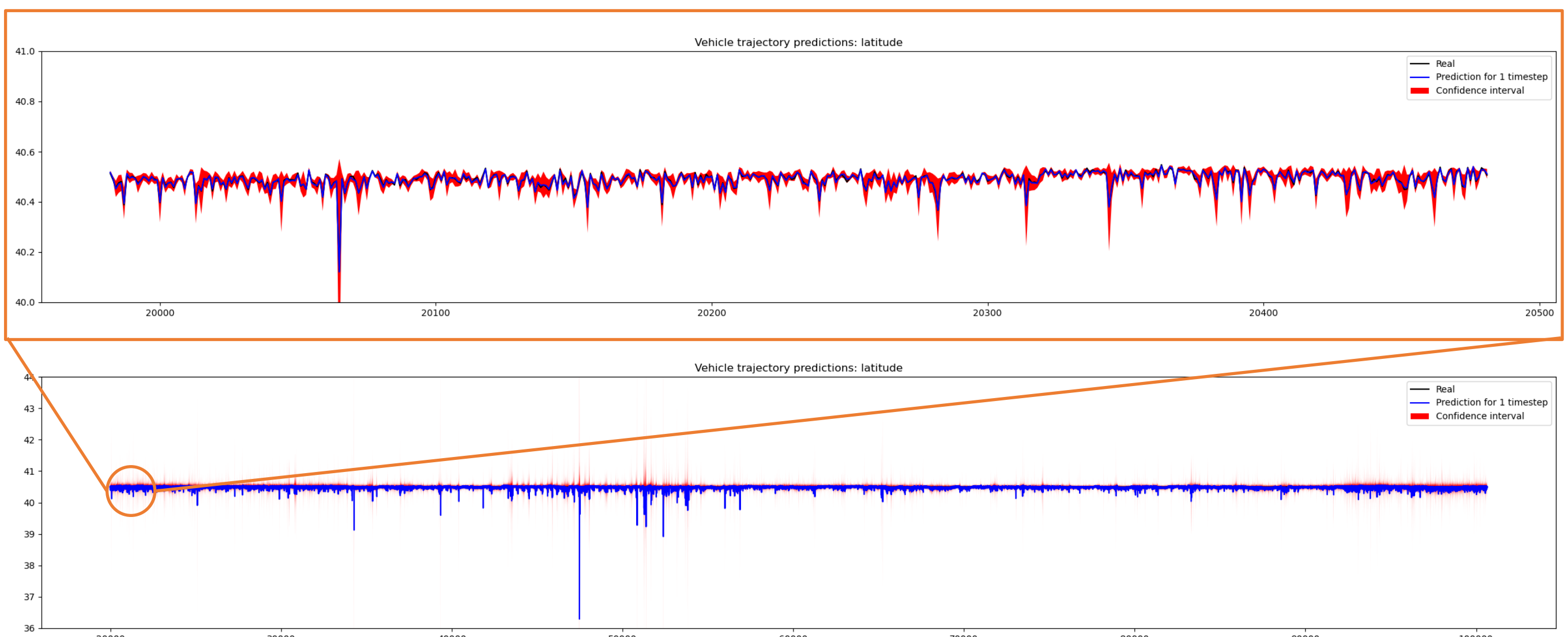

(b) Fine-tuned BLSTM model for latitude prediction

**Fig. 13.** Prediction results for fine-tuning the model with new cars.

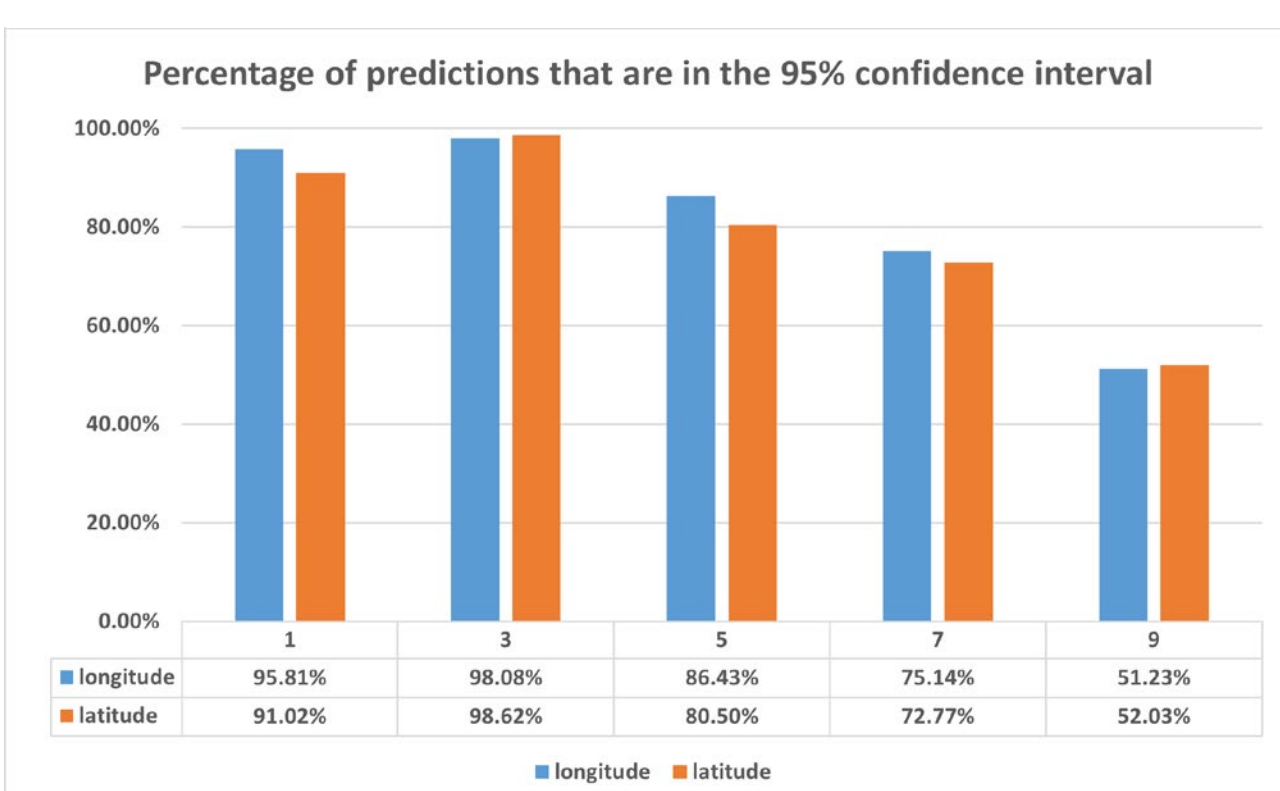

| | 1 | 3 | 5 | 7 | 9 |
|---|---|---|---|---|---|
| longitude | 95.81% | 98.08% | 86.43% | 75.14% | 51.23% |
| latitude | 91.02% | 98.62% | 80.50% | 72.77% | 52.03% |

**Fig. 14.** Prediction accuracy using last t=n frames.

Given the predicted location distributions of all vehicles at time step $t$, risk plots of the entire road segment can be generated. Figure 15 illustrates two vehicle position distributions at frames 9500 and 10000, respectively.

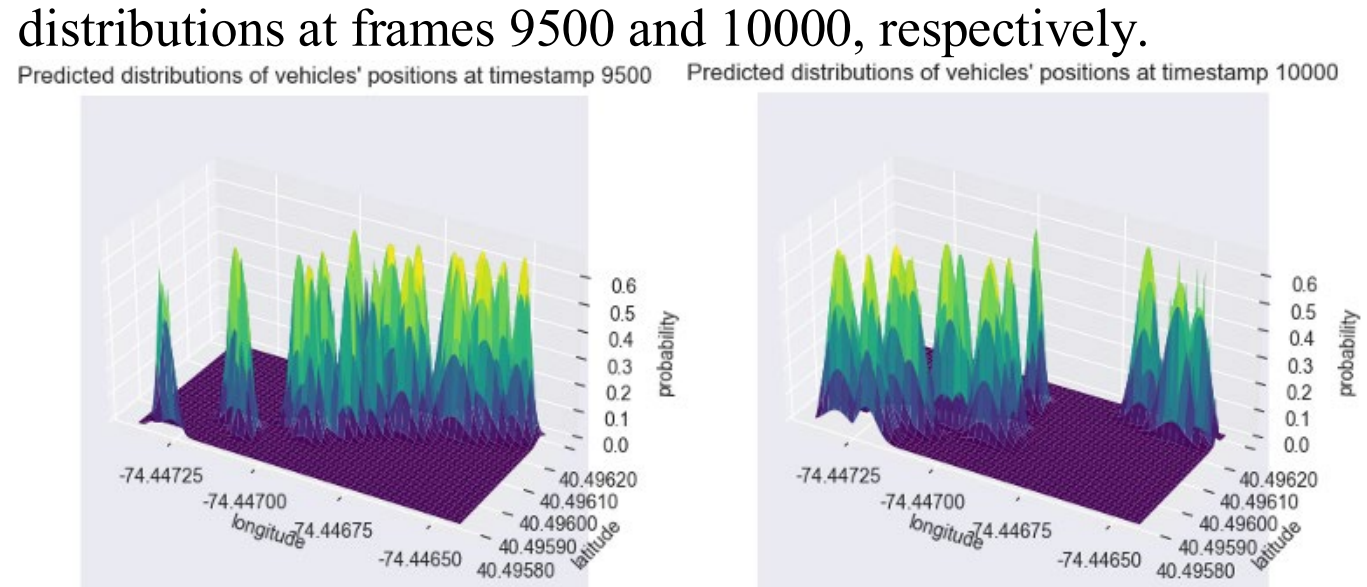

(a) Predicted location distributions of all vehicles at time step 9500

(b) Predicted location distributions of all vehicles at time step 10000

**Fig. 15.** Conflict risk plots of the road segment.

### *C. BLSTM-MPC Control Simulation*

In Section V-B, predictions were made based on video frames, with a resolution of 20 frames per second (FPS). This resolution is too high for real-world decision-making because human perception-reaction time is around 1.5 seconds and varies in different conditions [43]. The minimum action interval time for the BLSTM-MPC is set as 1 second to provide safety guidance for human drivers or to manipulate the automated vehicles. The risk scores are averaged over every 20 frames to generate the risk prediction for every second. Using the proposed hybrid automaton model described in Section IV-A, the risk for each state-space transition can be calculated from the conflict risk plot of the road segment. Figure 16 shows the conflict risk of each action if it is taken by the vehicle at each second. The predicted risks can be used as an advisory system for collision warnings. The driver will be alarmed if their driving behavior leads them to a dangerous situation. Also, a suggested action with a minor conflict risk is provided for the driver to assist in decision-making.

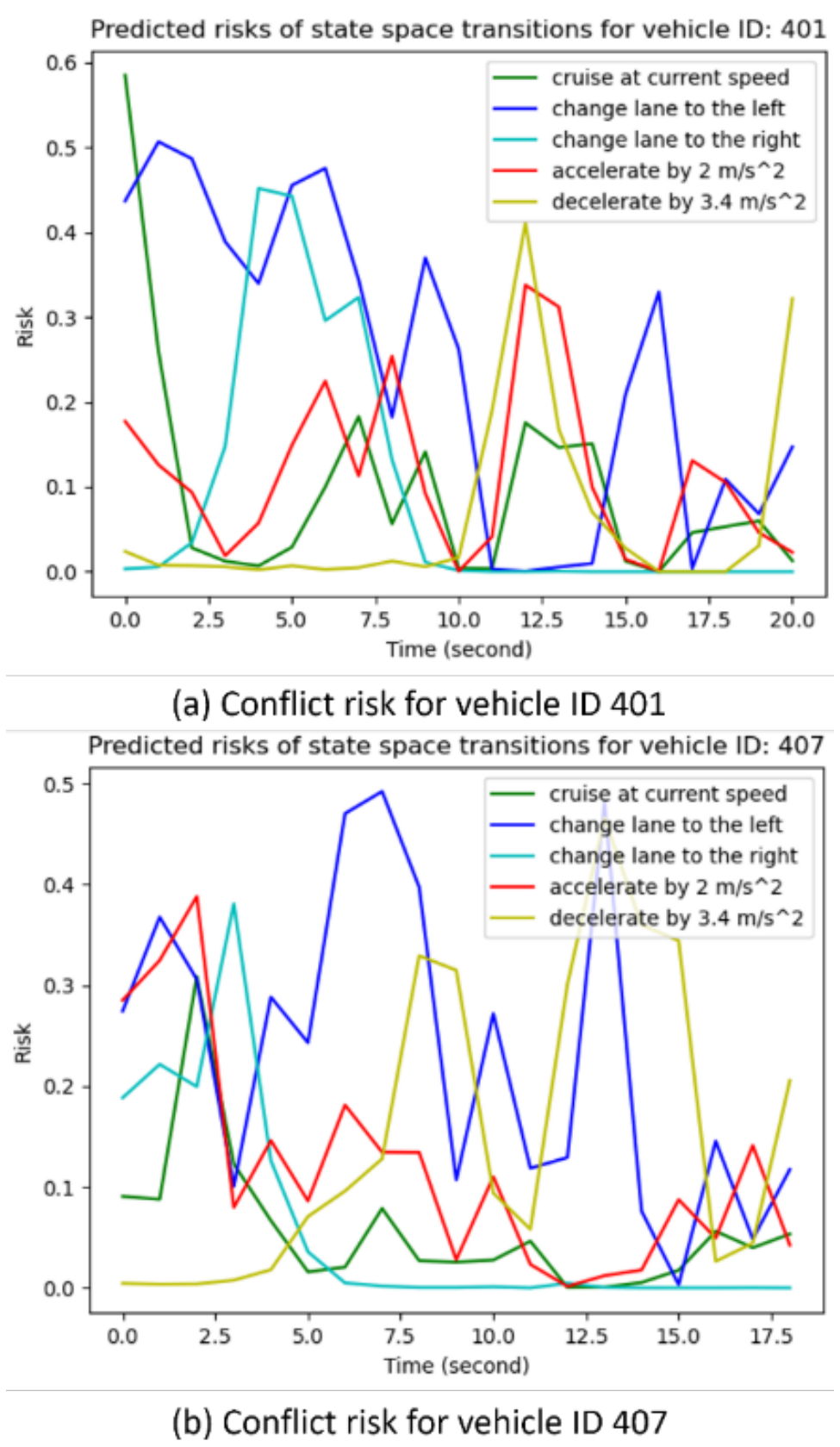

(a) Conflict risk for vehicle ID 401

(b) Conflict risk for vehicle ID 407

**Fig. 16.** Conflict risk for each action at each time step.

A driving simulation is created to test the performance of the BLSTM-MPC described in Section IV-B. Vehicles with ID numbers 401 and 407 from the collected data were selected to be the testing vehicles because they had the most surrounding vehicles in the collected dataset. The proposed controller manipulates them to demonstrate the model performance. Their driving environment is reproduced by the original data, which contains the initial position, speed of the testing vehicles, and the entire trajectories of all other cars. Figure 17 shows the simulation results, where the gray lines are the risks associated with the car maintaining its initial speed and staying in the same lane. These cruising cases are used as baselines for comparison between different control methods. The red lines are the actual risk scores of the cars under full human drivers' control. The yellow lines show the risk scores when the BLSTM-MPC algorithm controls the vehicles. The result indicates that the proposed hybrid predictive control method is able to choose a set of proper actions to lower the conflict risk compared to the human driver. Also, with the MPC looking forward to the next 5 seconds, it can take smoother control actions, thus avoiding sudden jumps in the conflict risk.

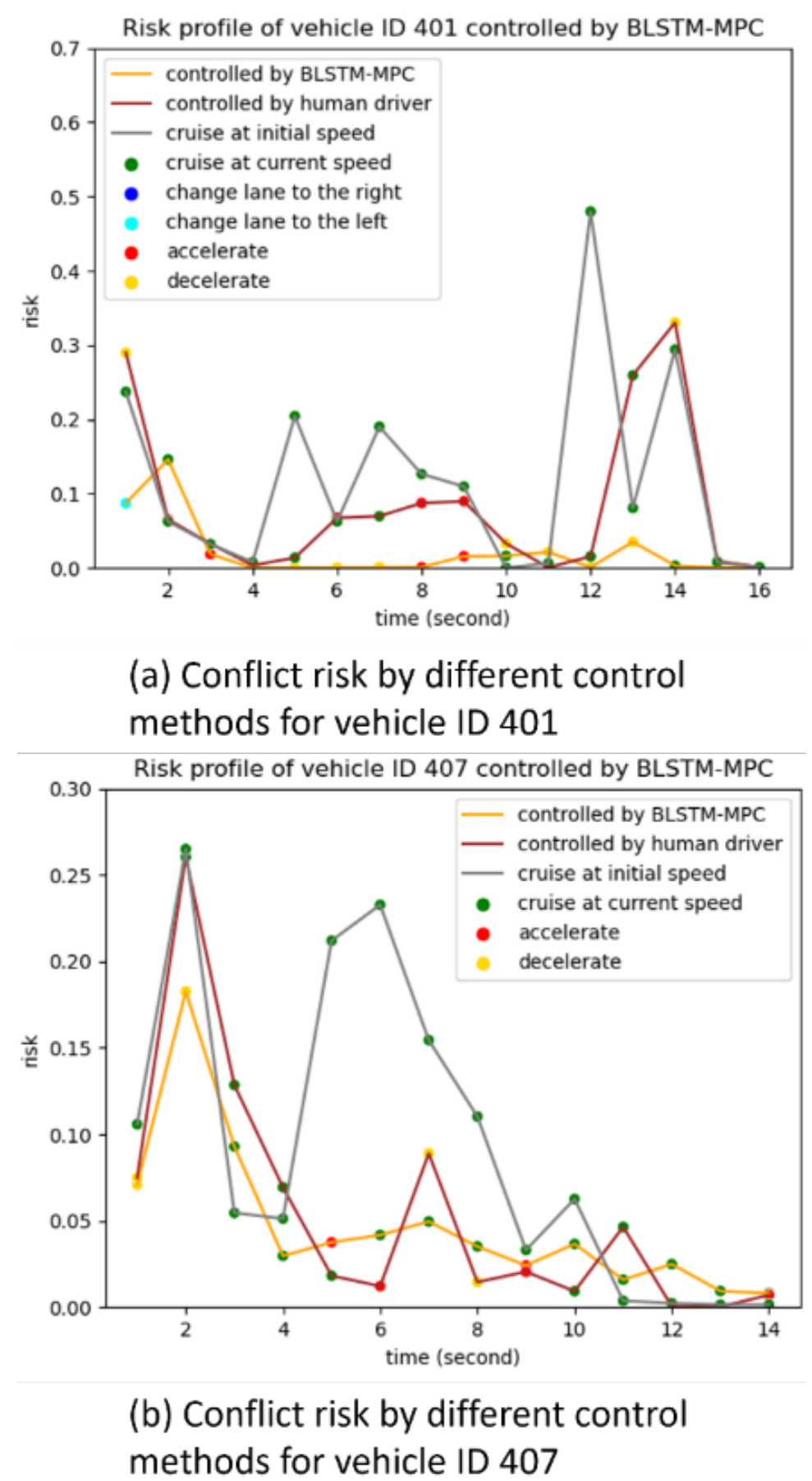

(a) Conflict risk by different control methods for vehicle ID 401

(b) Conflict risk by different control methods for vehicle ID 407

**Fig. 17.** Conflict risk comparison for each control method.

## VI. Conclusion

Using the V2I technique, this paper presents a probabilistic vehicle trajectory prediction approach by integrating LSTM with BNN to address the uncertain movement of vehicles. The testing result shows that the BLSTM model can effectively estimate the position distribution of vehicles in the next time step. The dynamic and probabilistic conflict risks are defined using the predicted distributions, and risk profiles are generated for vehicles' potential actions. Besides that, the authors proposed an idea that combines BLSTM with MPC for vehicle motion control, taking predicted conflict risks as the safety constraint. The neural network serves as a nonlinear risk prediction function and eliminates the limitation for MPC that the risk is hard to model in a closed-form expression. The experiment illustrates that the BLSTM-MPC algorithm can navigate vehicles through a safe path on the road. Finally, The BLSTM-MPC is tested and proved to perform better than human drivers in reducing traffic conflict risk.

There are certain restrictions in the work that we can improve in the future. For instance, only vehicles are detected from the roadside camera, and conflict risk predictions don't consider other obstacles and pedestrians. A more comprehensive image recognition method can be applied to detect road information. Also, the BLSTM-MPC models the driving maneuvers as a discrete control problem. The action space is pre-defined and limited to five basic motions and deterministic acceleration and deceleration rates. The driving situation is more complex in the real world, and intelligent vehicles may encounter more complicated cases, such as stopping at a stop sign or a red traffic light. The hybrid automaton can be extended, and more states and state-space transitions can be included.

## Appendix A

Input features of the BLSTM model:
(1) Lane ID: the id number of the lane the vehicle is driving on.
(2) Vehicle ID: a unique ID signed for each vehicle. Including vehicle ID in the model helps us differentiate individual driver behaviors.
(3) Current longitude of the vehicle (used as $x$ coordinate).
(4) Current latitude of the vehicle (used as $y$ coordinate).
(5) – (8) Dummy variables that indicate the geometry of the lane, whether it is a left-turn lane, a right-turn lane, a going-straight lane, or if the lane is going to merge with another lane. If all four variables are zeros, it means that the lane is going to diverge.
(9) An indicator of whether the vehicle is following (takes the value 1) or leading (takes the value 0).
(10) Headway distance.
(11) Speed difference to the front vehicle.
(12) Current speed.
(13) Current acceleration (positive value) or deceleration rate (negative value).
(14) Direction of the vehicle.
(15) The angle to the front vehicle.
(16) The speed of the front vehicle.
(17) The acceleration or deceleration rate of the front vehicle.
(18) The distance to the vehicle on the left.
(19) The angle to the vehicle on the left.
(20) The speed of the vehicle on the left.
(21) The distance to the vehicle on the right.
(22) The angle to the vehicle on the left.
(23) The speed of the vehicle on the right.
Note:
(15), (19), (22) will be zeros, and (16), (17), (18), (20), (21), (23) will be assigned an enormous value (e.g., 999) if there is no vehicle in that direction.

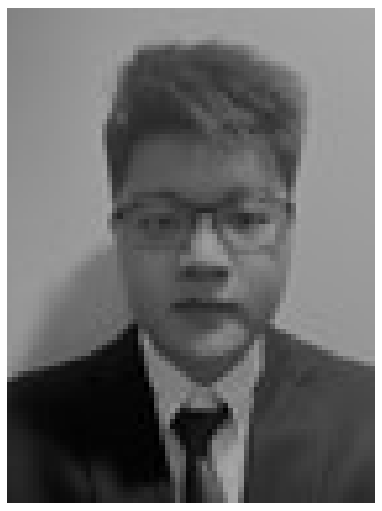

**Yufei Huang** received a B.Eng. degree in Automation from Xi’an Jiaotong University in 2016. Also, he received an MS degree in Systems Engineering from the University of Maryland, College Park in 2018. He is currently a Ph.D. student at Rutgers, the State University of New Jersey, studying Industrial and Systems Engineering, and a Research Assistant at the Center for Advanced Infrastructure and Transportation (CAIT). His research interests are in multi-agent systems, autonomous systems, robotics, and reinforcement learning.

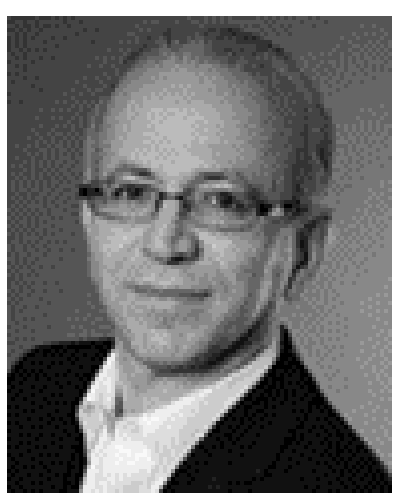

**Mohsen Jafari** (M’97) received a Ph.D. degree from Syracuse University in 1985. He has directed or co-directed a total of over 23 million US dollars in funding from various government agencies, including the National Science Foundation, the Department of Energy, the Office of Naval Research, the Defense Logistics Agency, the NJ Department of Transportation, FHWA, and industry in automation, system optimization, data modeling, information systems, and cyber risk analysis. He actively collaborates with universities and research institutes abroad. He has also been a Consultant to several Fortune 500 companies as well as local and state government agencies. He is currently a Professor and the Chair of Industrial & Systems Engineering at Rutgers University-New Brunswick. His research applications extend to manufacturing, transportation, healthcare, and energy systems. He is a member of the IIE. He received the IEEE Excellence Award in service and research.